\documentclass[10pt, conference, letterpaper]{IEEEtran}

\usepackage{amsmath}
\usepackage{amsthm}
\usepackage{amssymb}
\usepackage{ifthen}
\usepackage{graphicx}
\usepackage{subfigure}
\usepackage{color}
\usepackage{balance}
\usepackage{url}

\IEEEoverridecommandlockouts

\begin{document}

\title{Performance Analysis of Universal Robot Control \\ System Using Networked Predictive Control}

\author{\IEEEauthorblockN{Mahsa Noroozi\IEEEauthorrefmark{1} and Lorenz Kies\IEEEauthorrefmark{2}}
\IEEEauthorblockA{Institute of Communications Technology \\ Leibniz Universit\"{a}t Hannover \\ Hanover, Germany \\ Email: \IEEEauthorrefmark{1}mahsa.noroozi@ikt.uni-hannover.de,
\IEEEauthorrefmark{2}lorenz.kies@stud.uni-hannover.de}
}

\maketitle
\thispagestyle{plain} 
\pagestyle{plain}

\begin{abstract}
Networked control systems are feedback control systems with system components distributed at different locations connected through a communication network. Since the communication network is carried out through the internet and there are bandwidth and packet size limitations, network constraints appear. Some of these constraints are time delay and packet loss. These network limitations can degrade the performance and even destabilize the system. To overcome the adverse effect of these communication constraints, various approaches have been developed, among which a representative one is networked predictive control. This approach proposes a controller, which compensates for the network time delay and packet loss actively. This paper aims at implementing a networked predictive control system for controlling a robot arm through a computer network. The network delay is accounted for by a predictor, while the potential of packet loss is mitigated using redundant control packets. The results will show the stability of the system despite a high delay and a considerable packet loss. Additionally, improvements to previous networked predictive control systems will be suggested and an increase in performance can be shown. Lastly, the effects of different system and environment parameters on the control loop will be investigated.
\end{abstract}


\IEEEpeerreviewmaketitle

\section{Introduction}
\begin{figure}[h!]
    \begin{center}
    \includegraphics[width=0.85\linewidth]{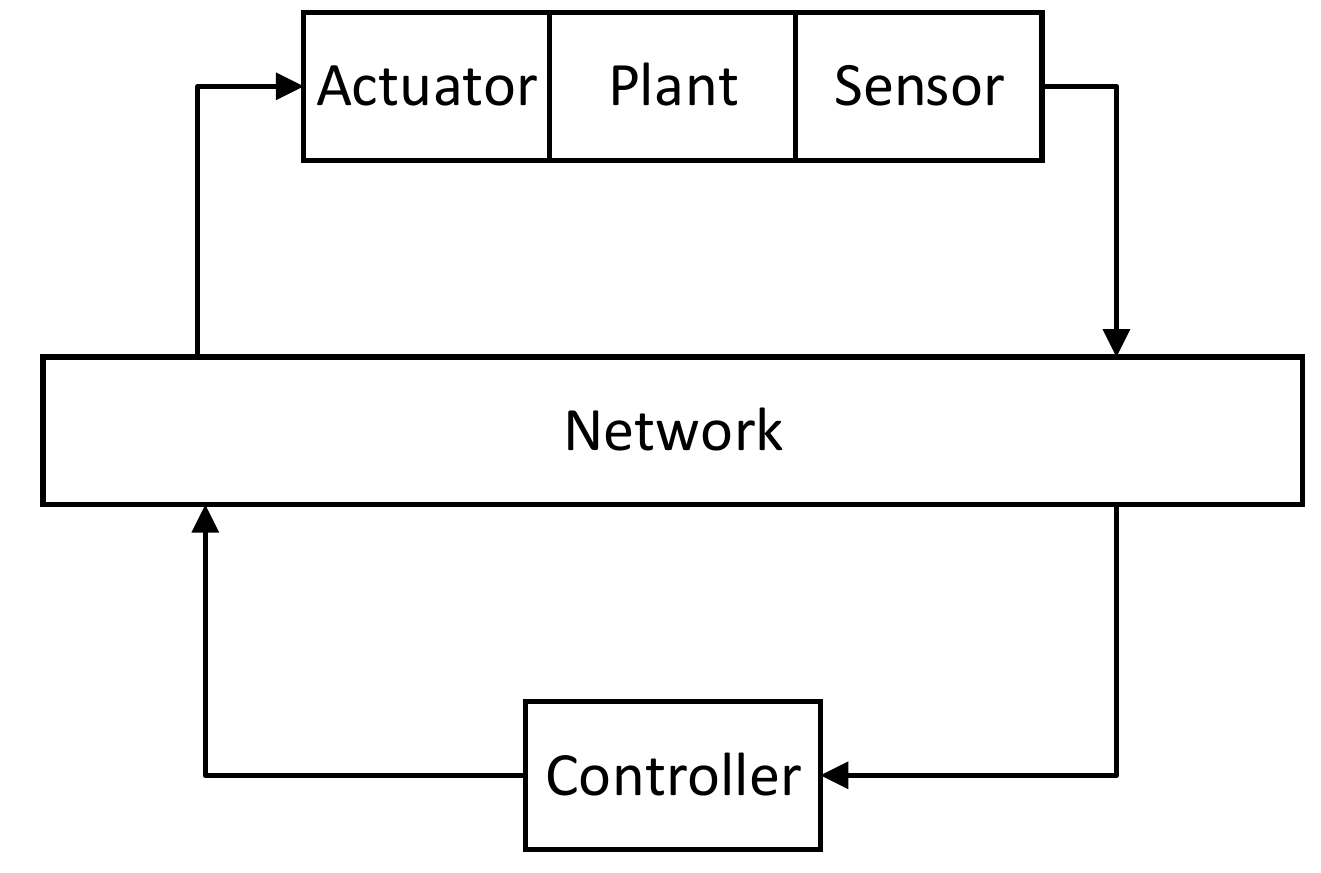}
    \caption{Basic structure of an NCS}
    \label{ncs.pic}
    \end{center}
\end{figure}

\IEEEPARstart{T}{he} use of a shared communication network to connect system components results in the concept of Networked Control Systems (NCSs). NCSs are feedback control systems containing plant, actuators, sensors, and controllers which may be located in geographically dispersed facilities. The general architecture of an NCS is shown in Fig. \ref{ncs.pic}. NCSs have been utilized in various areas, such as process control, industrial automation, vehicle industry, and robot manipulators \cite{Li,Zhang}.

Some advantages of using network architectures in NCSs are low installation and maintenance costs, reduced system wiring, and increased system flexibility. Even though the use of network architecture has many inherent benefits, it has some weaknesses due to the limited capacity of the network. As a result of traffic congestion in the network, there will be usually some constraints including network-induced delay and packet loss. Due to the existence of such constraints in communication network, the analysis and design of an NCS is complex. These constraints can also degrade the performance and even destabilize the system as well.

To mitigate the effects of network imperfections, diverse control methods including augmented deterministic discrete-time model \cite{halevi1988integrated}, queuing method \cite{luck1994experimental}, optimal stochastic control method \cite{nilsson1998real}, event-based method \cite{xi1998planning}, and predictive control \cite{liu2004networked} have been developed. Since periodically sending data at a high frequency can result in a heavy load on the network, \cite{zhang2020model} used an adaptive event-triggered controller to reduce network traffic. They also included measurement uncertainties in their model and introduced dynamic output feedback control to demonstrate the performance. To compensate some of the constraints in NCS design, Networked Predictive Control (NPC) has attracted more and more attention. Networked predictive control systems are NCSs that use predictive controllers as an active compensation method for network communication constraints. \cite{zhang2012design} studied the design, simulation, and implementation of networked predictive control.

Delays occurring in the forward and feedback channels in NCSs are distinguished apart. Regarding the forward channel, the controller doesn't know the time taking the control signal to reach the actuator from controller. Thus an exact correction can't be achieved while calculating the control signals. To measure the delays of the end-to-end communication, Round Trip Time (RTT) delay can be used. RTT in NCSs is used to measure the total time delay of a control cycle. Instead of considering forward and feedback delay separately, \cite{li2018robust,vafamand2018networked,pang2021active} combined it to the RTT delay and based their prediction and delay compensation on this measure. This eliminated the need for synchronization between the two sites. \cite{chen2011stability} used NPC to teleoperate a robot with significant variable RTT delay. They proposed a sparse multi variable linear regression algorithm to predict the RTT delay with good accuracy.

In a lot of earlier work, the delay was only in the order of a few sample times of the system and rarely more than the system dynamics. All in all a lot is to be gained from the advantages of NPC, making it a worthwhile effort to develop tools for their design \cite{zhang2015survey,mahmoud2016networked}. A networked predictive control system will be here implemented and variations in parameters and architecture will be compared. The base architecture consists of a predictor, controller, forward channel, delay compensator, plant, and feedback channel. The system to be controlled is a universal robot arm.

The remainder of this paper is organized as follows. Firstly, an overview of formal description, the structure of the NPC approach, and the implementation of the system are given in section \ref{sec2}. In section \ref{sec3}, the evaluation results with different features or parameters are shown. Finally, section \ref{sec4} concludes the paper.

\section{Networked predictive control (NPC)}\label{sec2}
\begin{figure}[htb]
    \begin{center}
    \includegraphics[width=0.95\linewidth]{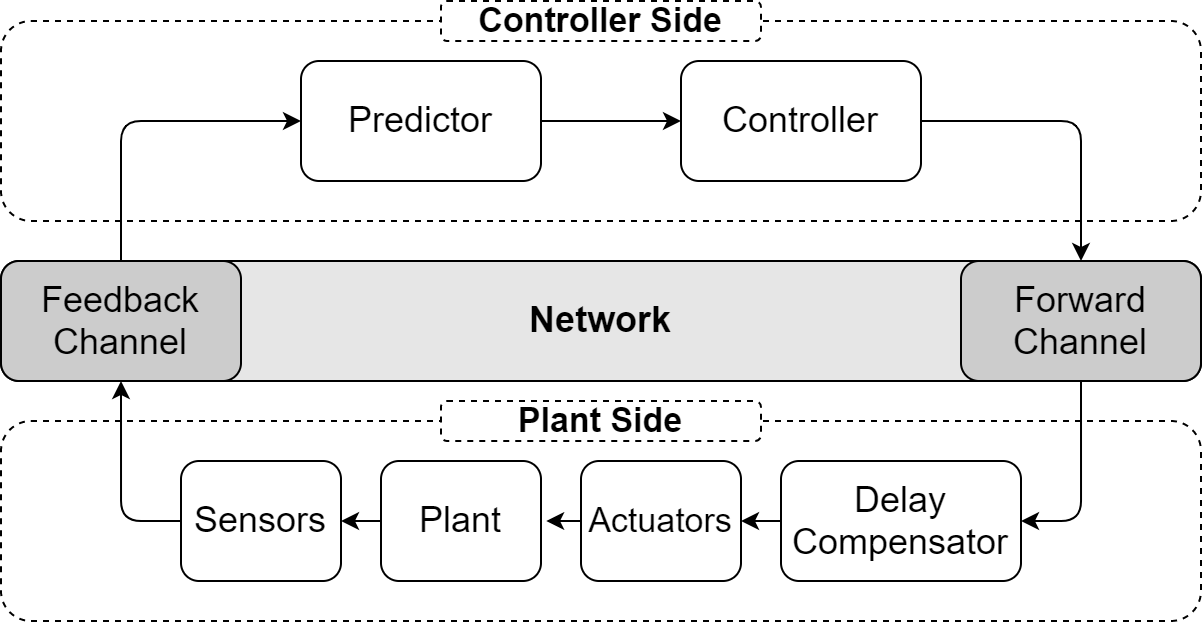}
    \caption{Architecture for a Networked Predictive Control System}
    \label{architecture.pic}
    \end{center}
\end{figure}

Networked Predictive Control Systems (NPCSs) are control systems where the control loop is closed through a network with some extra components compared to a regular control loop. The basic structure of an NPCS can be seen in Fig. \ref{architecture.pic}. The basic setup shares some similarities with a regular control loop like the one shown in Fig. \ref{ncs.pic}, but there are a few additional features namely predictor, delay compensator, and network, which encompasses the forward and feedback channel.

\subsection{Plant model}\label{section:plant}
The plant used is a 6-axis cooperative universal robot arm. The inverse kinematics are handled on site and the robot is controlled by acceleration in all three spacial dimensions. Since all of these dimensions should behave the same, only one is used to simplify the model. The resulting system description is: 
\begin{equation}
\centering
    \dot{\boldsymbol{x}}(t) = \boldsymbol{f}(\boldsymbol{x}(t),\boldsymbol{u}(t))
    = \begin{bmatrix}\dot{x}_1(t)\\\dot{x}_2(t)\end{bmatrix} = \begin{bmatrix}x_2(t)\\u_1(t)\end{bmatrix}
    \label{1}
\end{equation}
or as a linear system in state space description:
\begin{equation}
\centering
    \dot{\boldsymbol{x}}(t) =                                \boldsymbol{A}\boldsymbol{x}(t)+\boldsymbol{B}\boldsymbol{u}(t)
    = \begin{bmatrix}0 & 1\\0 & 0\end{bmatrix}\boldsymbol{x}(t)
    + \begin{bmatrix}0 \\ 1 \end{bmatrix}\boldsymbol{u}(t)
    \label{2}
\end{equation}
with $\boldsymbol{x}(t)$ as the state consisting of the position and velocity, and $\boldsymbol{u}(t)$ as the control input. The output of the system is:
\begin{equation}
\centering
    \boldsymbol{y}(t) = \boldsymbol{g}(\boldsymbol{x}(t),\boldsymbol{u}(t))
    = \begin{bmatrix}y_1(t)\\y_2(t)\end{bmatrix}
    = \begin{bmatrix}x_1(t)\\x_2(t)\end{bmatrix}
\end{equation}

The discrete sample time $T$ or the sample time of the robot is $8{ms}$. To provide a more challenging environment that can better represent the problems of a real system, Gaussian noise is added to the measurement. The noise is generated by adding a random value that is chosen from a Gaussian normal distribution to both of the output vector elements.

It is a very conservative assumption, as real sensors for the position and speed of the robot arm would be more precise, the position readout from the universal robot is accurate to $0.03 mm$ \cite{pollak2020measurement}.

\subsection{Controller}\label{section:control_law}
A Linear Quadratic Regulator (LQR) is used in this work as a controller. LQR is a linear full state feedback controller, that is optimal in its solution space determined by weights for errors in different parts of the state and costs for control inputs. It is a state space description but without the output equation:

\begin{equation}
\centering
    \dot{\boldsymbol{x}}(t) = \boldsymbol{A}\boldsymbol{x}(t)+\boldsymbol{B}\boldsymbol{u}(t)
\end{equation}
\begin{equation}
\centering
    \boldsymbol{u}(t) = \boldsymbol{K}\cdot\boldsymbol{x}(t)
\end{equation}
with $\boldsymbol{K}$ as a gain vector or matrix for the case of a multiple input system.

The cost function that the controller minimizes is:
\begin{equation}
\centering
    J(\boldsymbol{Q},\boldsymbol{R}) = \int \boldsymbol{x}(t)^T\boldsymbol{Q}\boldsymbol{x}(t)+ \boldsymbol{u(t)}^T\boldsymbol{R}\boldsymbol{u(t)} \,dt
\end{equation}
where $\boldsymbol{Q}$ is the weight matrix for the different parts of the state and $\boldsymbol{R}$ is the weights for the control channels \cite{shaiju2008formulas}.

The weight matrices and vectors were chosen as follows, $\boldsymbol{A}$ and $\boldsymbol{B}$ are the same as in \ref{2}.
\begin{equation}
\centering
    \boldsymbol{Q} = \text{diag}(1, 0.5)\quad\text{and}\quad R = 0.1
\end{equation}
This results in the following gain vector $\boldsymbol{K}$:
\begin{equation}
\centering
    \boldsymbol{K}^T = [3.1623, 3.3652]
\end{equation}
were calculated using the Python control systems library \cite{fuller2021python}.

\subsection{Network}\label{Network}
The network is modeled using the forward and feedback channel. These can be different physical channels or conceptual channels that are run through a shared communication network like a bus system or a computer network. The properties of these channels create the need for the predictor and the delay compensator.

\begin{figure}[htb]
    \begin{center}
    \includegraphics[width=0.90\linewidth]{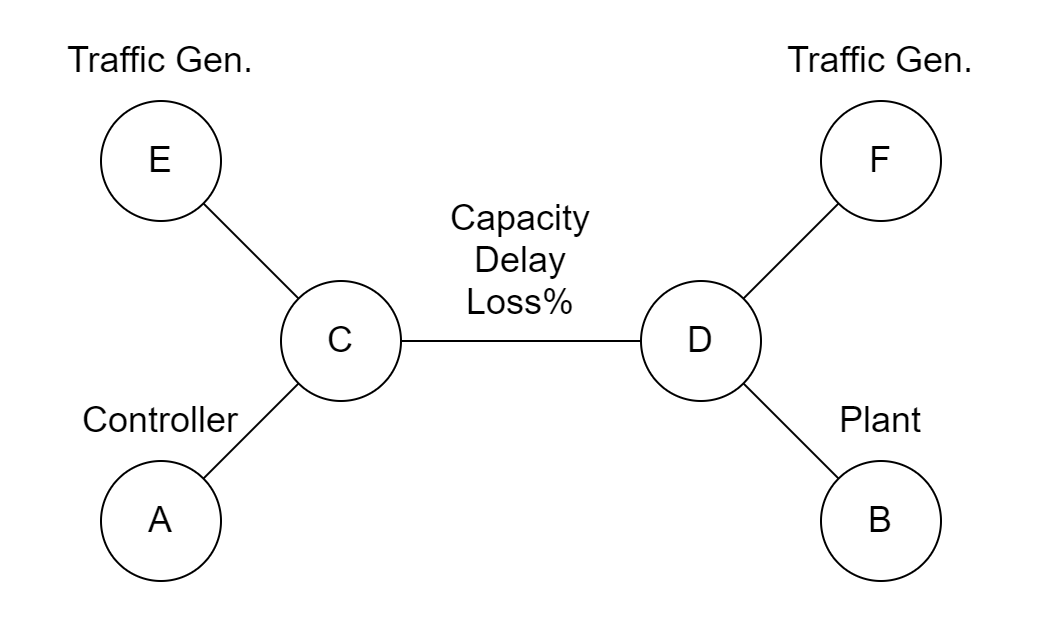}
    \caption{Network topology of the experiment}
    \label{handlebar_network.pic}
    \end{center}
\end{figure}

The simulation was done using a network emulator. The test network is made up of 6 nodes as shown in Fig. \ref{handlebar_network.pic}. All traffic in NPCS has to pass through the bottleneck between node $C$ and $D$ which has a limited capacity, a transmission delay, and probably packet loss.

Hosts $E$ and $F$ are sending dummy traffic to each other with similar characteristics to real internet traffic. This traffic uses part of the link capacity between $C$ and $D$ causing packets from the plant and controller to be lost or delayed.

\subsubsection{Delay functions} \label{subsection:delay_functions}
The function to model the delay of the communication channel is a sum of sine waves with random phase shifts. The random values are generated from a supplied seed for repeatability. The expression for the function is the following:
\begin{equation}
\centering
    d(t) = a_0+\sum_{i=0}^{n-1} \sin{(1.2^i\cdot f_0\cdot 2\pi\cdot t+\varphi_i)}\cdot \frac{c}{1.2^i}
\end{equation}
with $c$ as a normalisation factor to bring the maximum deviation with respect to $a_0$ to the desired deviation $d$. The factor is calculated as follows:
\begin{equation}
\centering
c = d\cdot 1.77\cdot\left({\sum_{i=0}^{n-1} \frac{1}{1.2^n\cdot}}\right)^{-1}
\end{equation}

With the decreasing amplitudes of the higher frequencies, the function has a fractal like behaviour, where the amount of higher frequencies only increases the detail but not the general shape.

\subsection{Predictor}\label{Predictor}
Due to the delay introduced by the network, the control values that arrive will be out of date. The predictor uses a model of the plant to predict future states. It needs to know the initial conditions through the feedback channel and all inputs that will be applied to the plant up to the desired prediction point, which it obtains from the control buffer. With this information it can integrate the system's differential equations and calculate the new state.

With increasing network delay, the predictor has to integrate farther into the future. This means, that especially for longer delays, it is very important to have an accurate model of the plant. To achieve this, it runs an online parameter estimator that continually updates the parameters of the plant model by observing how the plant state evolves over time and how it reacts to control inputs. The result of this predictor controller cycle is a series of plant states and control values.

\label{control_value_reuse}
Another way to improve the performance of the predictor is to use part of the control values of the last iteration. To implement this, the controller needs to know the RTT. So it knows the control values that the controller will have used and control values that can be improved upon with newer information. Since the control unit always predicts and calculates all states and control inputs from zero delay up to the prediction horizon, the control packet will get quite large. This is a waste of bandwidth as the plant should only need one of those values or rarely a few more if other control packets are lost. To mitigate this, the control unit sends a section of the series with the values that are most likely to get used. To know these values, it needs to know the packet arrival time, which can be calculated using RTT. Then, it keeps a small window of values around that time point, while discarding the beginning and end of the series.

\subsection{Delay Compensator}\label{Compensator}
Since the control unit's prediction of when the control information will arrive at the plant may be wrong, it can calculate a sequence of control values for different arrival times. The delay compensator takes this sequence of control values, buffers, and chooses the appropriate control value from the most recent packet and apply it to the plant. Every control packet contains a series of control values spanning a small time window around the predicted arrival time. When packets are lost, the delay compensator uses the following values of the series of the last control packet it received. If it reaches the end of the series, it will hold the last value until it receives a new packet.

\section{Results} \label{sec3}
In this section, we evaluate and verify the proposed design. We compare either the different features or variations of a parameter within a range. A large part of the development and experiments were done as offline simulations as opposed to online real time simulations.

\subsection{Metrics}
The parameters of the control unit are following:

\begin{table}[h!]
\centering
\begin{tabular}{| c | c |} 
 \hline
 timeStep & 0.008 s  \\ 
 \hline
 controllerHorizon & 100 \\ 
 \hline
 delayMargin & 16 \\
 \hline
 useIntegrator & True \\
 \hline
 useOldControl & True \\
 \hline
 useEstimator & True \\ 
 \hline
\end{tabular}
\end{table}
The time step is the sample time of the plant and controller, the controller horizon is the prediction horizon of the control unit, and the delay margin is the length of a sequence of control values. The boolean variables enable or disable the respective features. The combination of controller horizon and time step results in a total of $0.8 s$ look ahead time. This is greater than the maximum RTT that can reasonably be expected.

\subsubsection{Offline experiments}
Different parameters are used for the delay functions. The base frequency is $0.4 Hz$, base delay is equal to $125 ms$, delay variation is set to $100 ms$, and there are $20$ sub-frequencies. It means that the average RTT is $250 ms$ and the propagation time for fiber optics is around $66 ms$.

\subsubsection{Real time experiments}
While the offline simulations enabled quick developments and were generally easier to execute, they are farther from reality and therefore yielded less meaningful results. The discrete event simulation was replaced here by a real time simulation running on a simulated network. The experiments were essentially the same as the offline simulations with a few network specific additions. The base parameters of the bottleneck between $C$ and $D$ were the capacity of $25 Mbps$, delay of $125 ms$, and loss of $0 \%$.

\subsection{Robustness and Performance}
The results are evaluated by comparing them to a local delay free controller, which may have deviations due to a bad plant model, measuring uncertainties, or inappropriate controller gains. The comparison will only consist of errors due to the network and network error compensation. The deviation is accumulated to a scalar value by taking the sum of the squares of the difference between ideal and actual value. This sum is then divided by the total evaluated time to take experiment run time out of the equation:
\begin{equation}
\centering
    RSS = \frac{1}{n}\cdot\sum_{i=0}^{n-1}(x_{1,ideal}(i)-x_1(i))^2
\end{equation}
$n$ is the number of samples, $x_{1,ideal}(k)$ is the ideal state resulting from local control and $x_1(k)$ is the actual state. Only the first part of the state, the position, was evaluated.

\subsubsection{Parameter Estimator}
\begin{figure}[htb]
\centering
\subfigure[Offline]{
\includegraphics[width=0.95\linewidth]{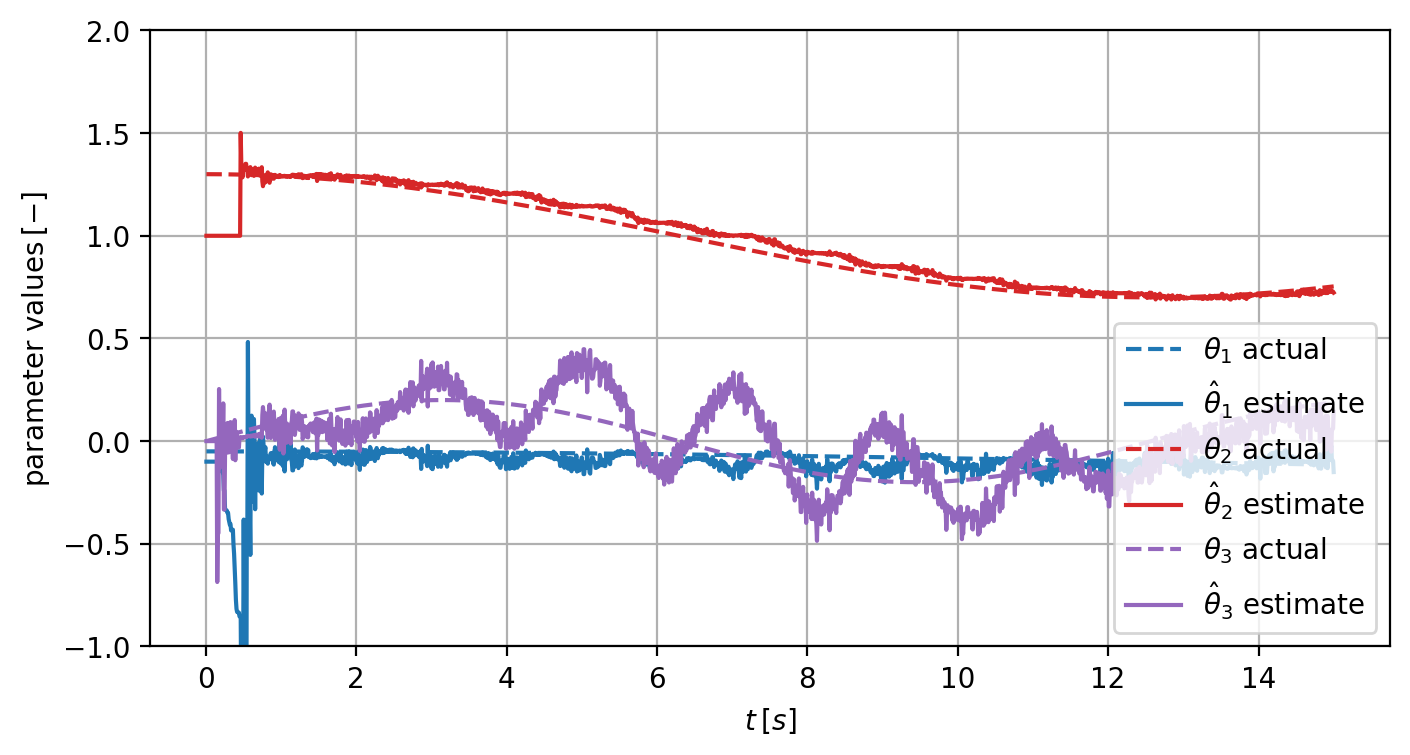}
\label{fig:estimator1}
}
\subfigure[Real time]{
\includegraphics[width=0.95\linewidth]{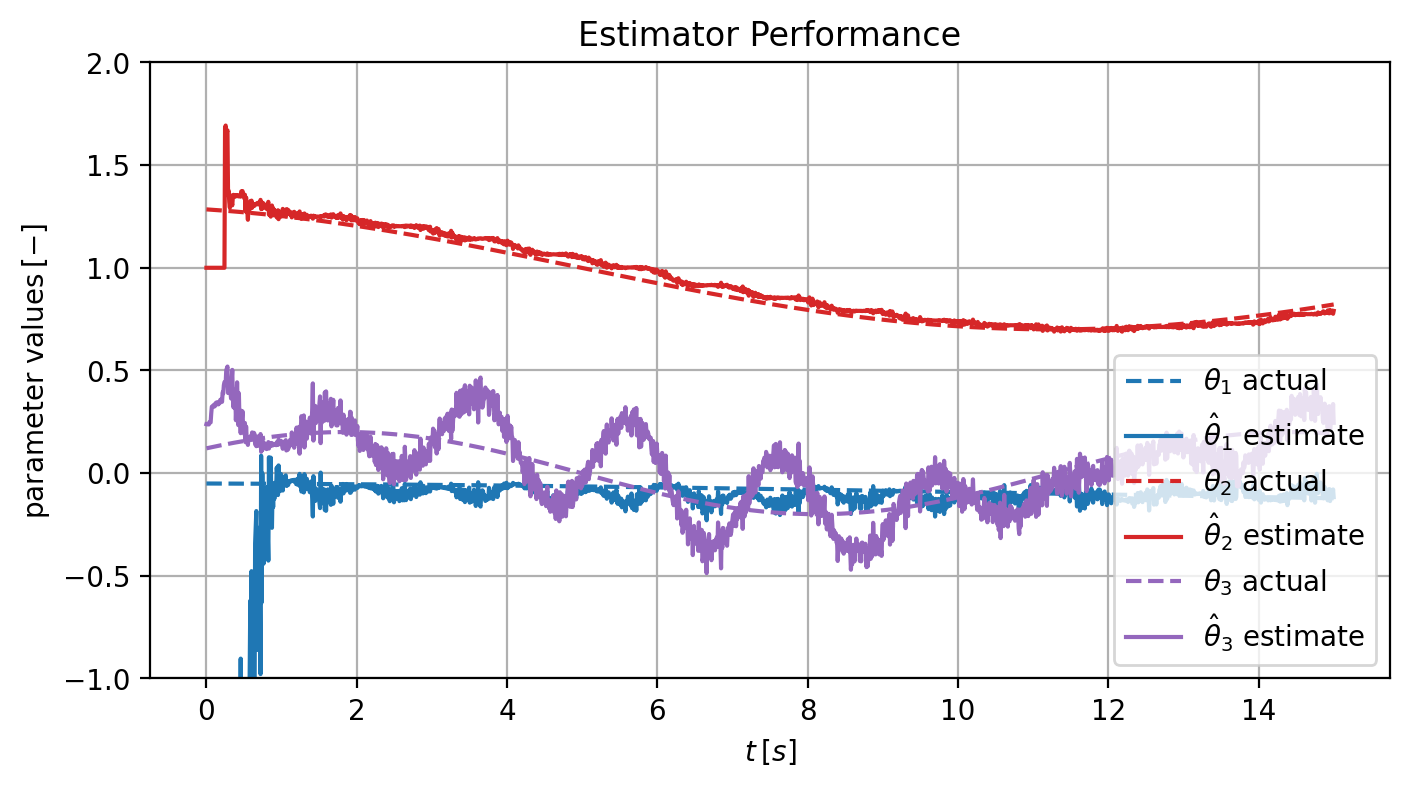}
\label{fig:estimator2}
}
\caption{Parameter tracking of the estimator}
\label{fig:estimator}
\end{figure}

Fig. \ref{fig:estimator} shows the performance of the parameter estimator. As the figure shows, the estimator tracks the changing parameters with reasonable accuracy. All of the estimates show noticeable noise, this is due to the measurement noise. The effect of noise could be reduced by increasing the forgetting factor $\lambda$ to give more weight to older values, however this would also cause the estimator to react slower to changes in parameters.

The estimates for $\theta_2$ and $\theta_3$ show oscillations that resemble the reference frequency. This is because the estimator has trouble differentiating between the different sources that influence the prediction expression. Similarly to the estimator following the noise, this can also be mitigated by increasing $\lambda$, but would increase the reaction time.

In general, the performance for each parameter seems to scale with the magnitude of the parameter's effect on the system and the proximity of the shared forgetting factor to the optimal individual forgetting factor. Despite these shortcomings, the estimator still improves the performance of the control unit, as the estimated parameters are generally closer to the actual values than the constant approximations without the estimator.

The estimator performance in real time experiments was also very similar to that of the offline simulations, as can be seen in Fig. \ref{fig:estimator2}.

\subsubsection{Using the Predictor}
\begin{figure}[htb]
\centering
\subfigure[Offline]{
\includegraphics[width=0.90\linewidth]{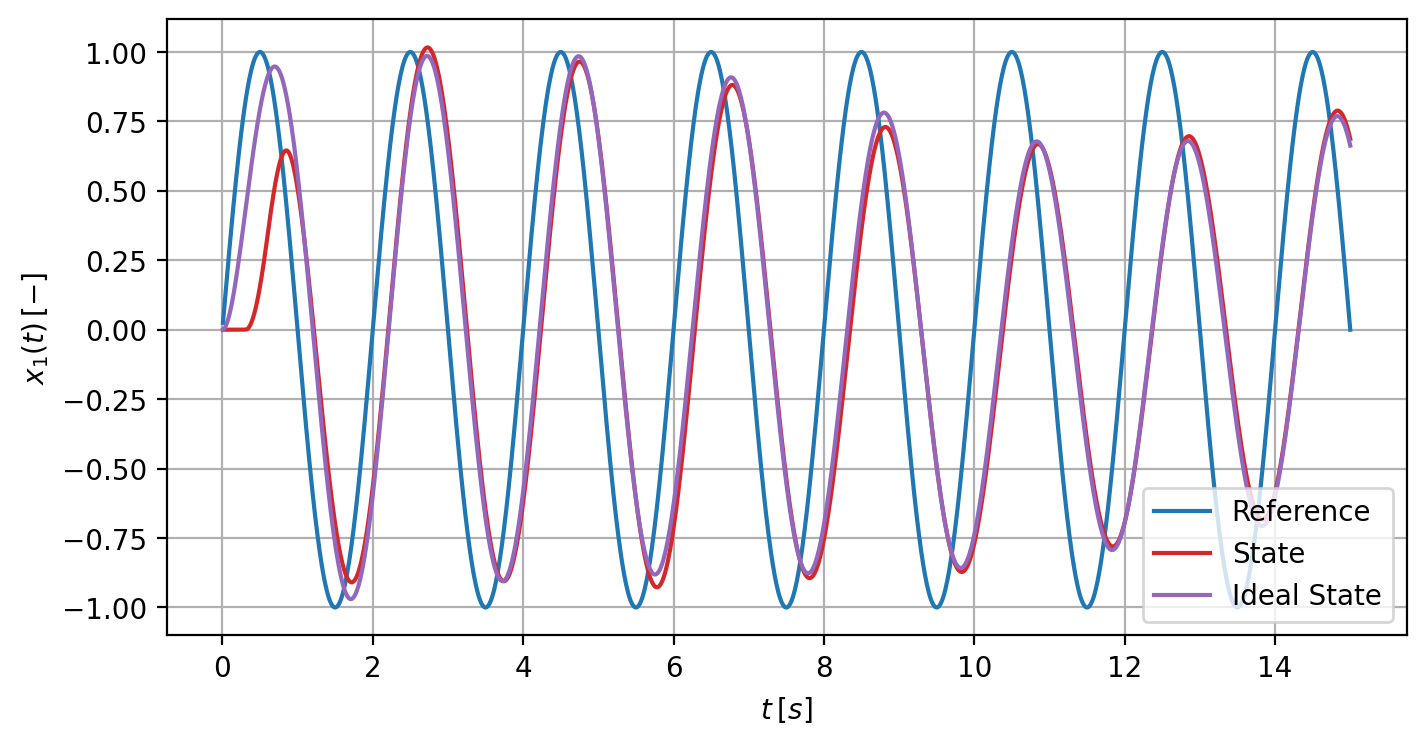}
\label{fig:predictor1}
}
\subfigure[Real time]{
\includegraphics[width=0.90\linewidth]{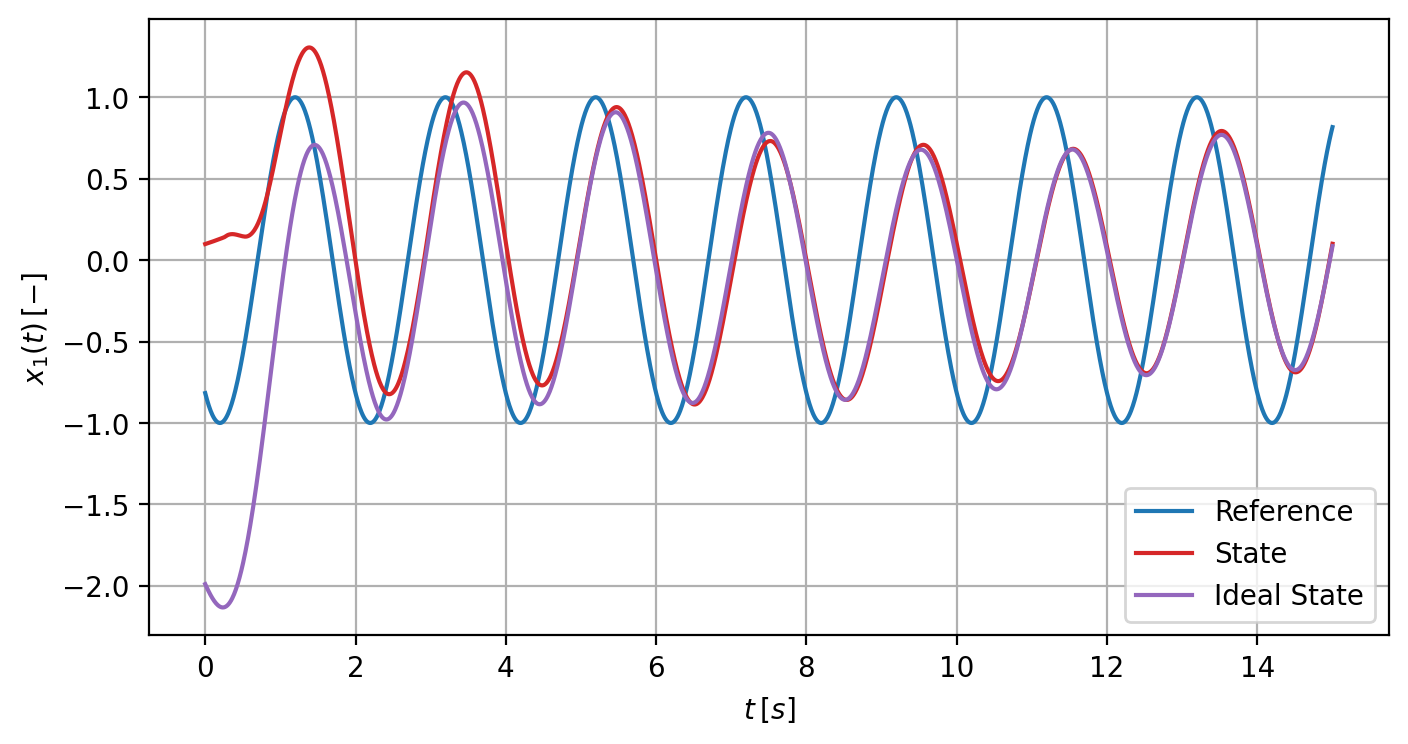}
\label{fig:predictor2}
}
\caption{State vs ideal state over time using predictor}
\label{fig:predictor}
\end{figure}

Fig. \ref{fig:predictor} shows the reference tracking compared to the ideal delay free system. At the start, the error is very noticeable, as the controller will only send control values after it has received the first system state. This means that the system will do nothing for one RTT, so that the first $5$ seconds are ignored in performance evaluation. It can also be seen that the ideal state does not track the reference or even an amplitude and phase shifted version of the reference. This is because the controller for the ideal system is not designed to compensate for some errors of the plant including the errors resulting from the change of parameters. It doesn't implement additional features to actually track the reference. But a very good performance can be seen as the actual state behaves very close to the delay free ideal.

The real time version was able to show similarly promising results as the offline version. The RSS value was only about twice that of the offline version, which is still a good result. Fig. \ref{fig:predictor2} shows the first $15$ seconds of the experiment, showing a similar performance to the offline equivalent Fig. \ref{fig:predictor1}.

\begin{figure}[htb]
\begin{center}
\includegraphics[width=0.90\linewidth]{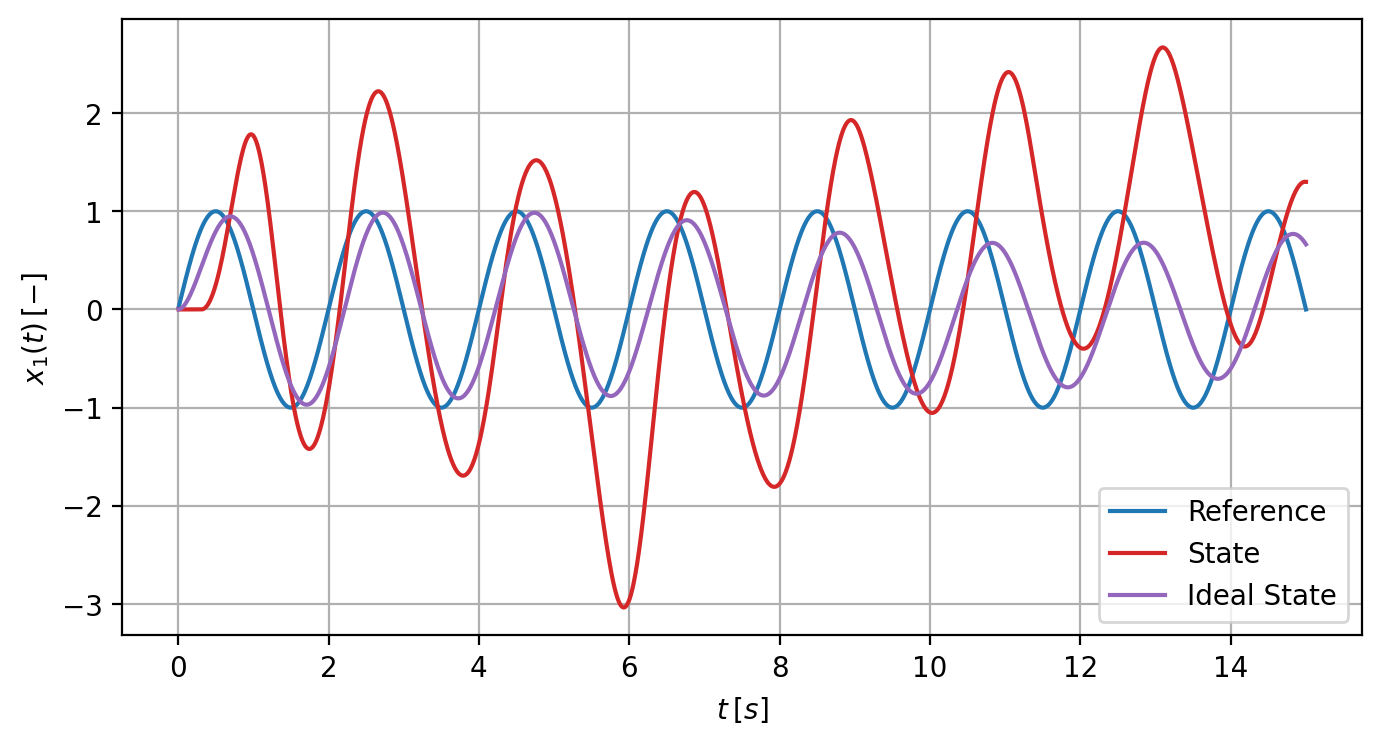}
\caption{State vs ideal state over time without predictor}
\label{nopred}
\end{center}
\end{figure}

Without the predictor, the control unit effectively assumes that there is no delay. The performance of the system without predictor can be seen in Fig. \ref{nopred}. It is evident that the controls do not lead to the system following the reference and the system seems unstable.

\subsubsection{Performance over Varying Average Delay}
\begin{figure}[htb]
\centering
\includegraphics[width=0.85\linewidth]{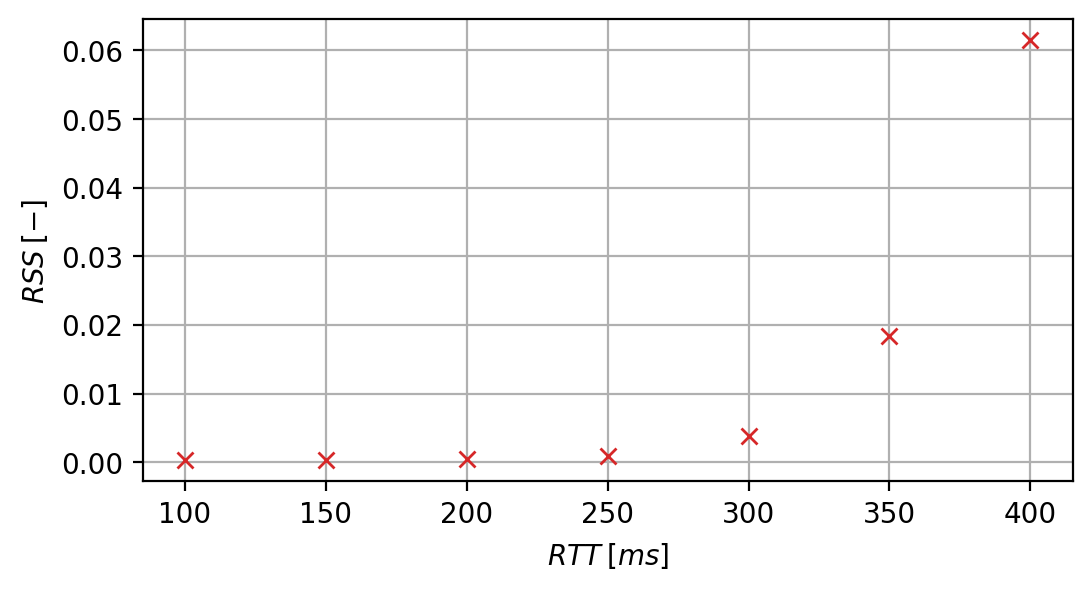}
\caption{Performance over varying average delays}
\label{fig:delay}
\end{figure}

Another effect to examine was the impact of different RTTs on performance. Fig. \ref{fig:delay} shows the effect of the average RTT on the performance. Note that in the offline simulation the change of the variation was proportional to the average delay. As expected, the error increases with increasing delay, and it does so quite drastically at higher delays. This is to be expected, as inaccuracies in the initial state or parameters can compound exponentially.

The real time experiment was slightly different to the offline version. In the real time version, only the base delay of the bottleneck between $C$ and $D$ was set to the delay values of $[50, 75, 100, 125, 150, 175, 200] ms$. The total delay however is dependent on the delay that was resulted from buffer times and other network phenomena. This means that the actual RTT was slightly higher. In a similar trend to the previous experiments, the qualitative picture is the same as the offline version but with worse performance in general.

\subsubsection{Variation of Delay Margin}
\begin{figure}[htb]
\centering
\includegraphics[width=0.85\linewidth]{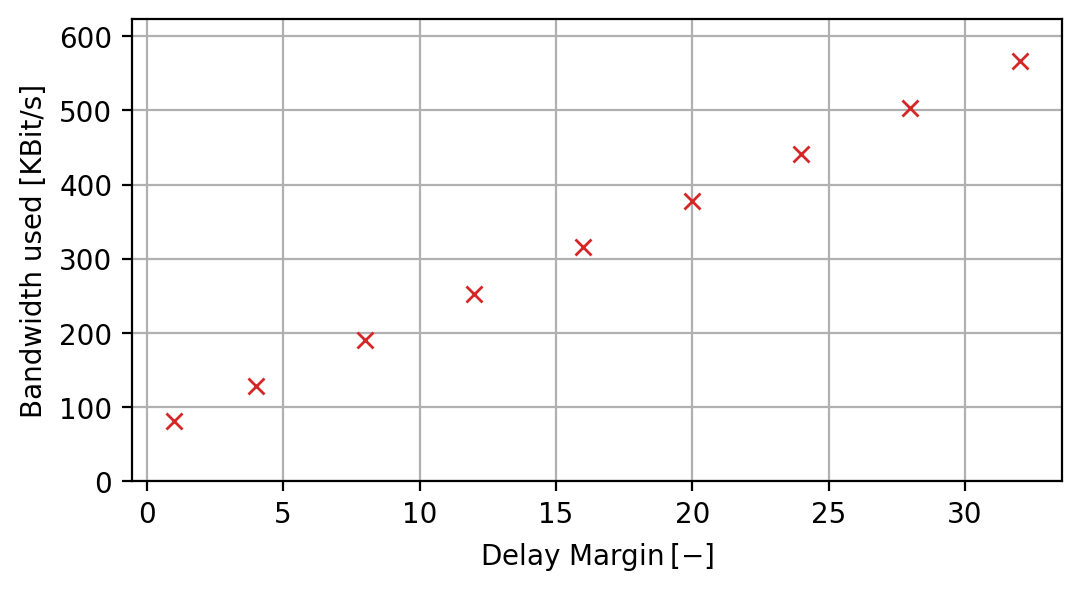}
\label{fig:margin1}
\caption{Real time bandwidth usage}
\label{fig:bandwidth}
\end{figure}

To investigate the hypothesised trade off between performance and bandwidth when changing the delay margin, the system performance and outgoing bandwidth of the controller were measured with different delay margins. As shown in Fig. \ref{fig:bandwidth}, the bandwidth grows linearly with the delay margin, since the individual packet size is linearly dependent on the delay margin. 

\begin{figure}[htb]
\centering
\subfigure[Real time performance]{
\includegraphics[width=0.80\linewidth]{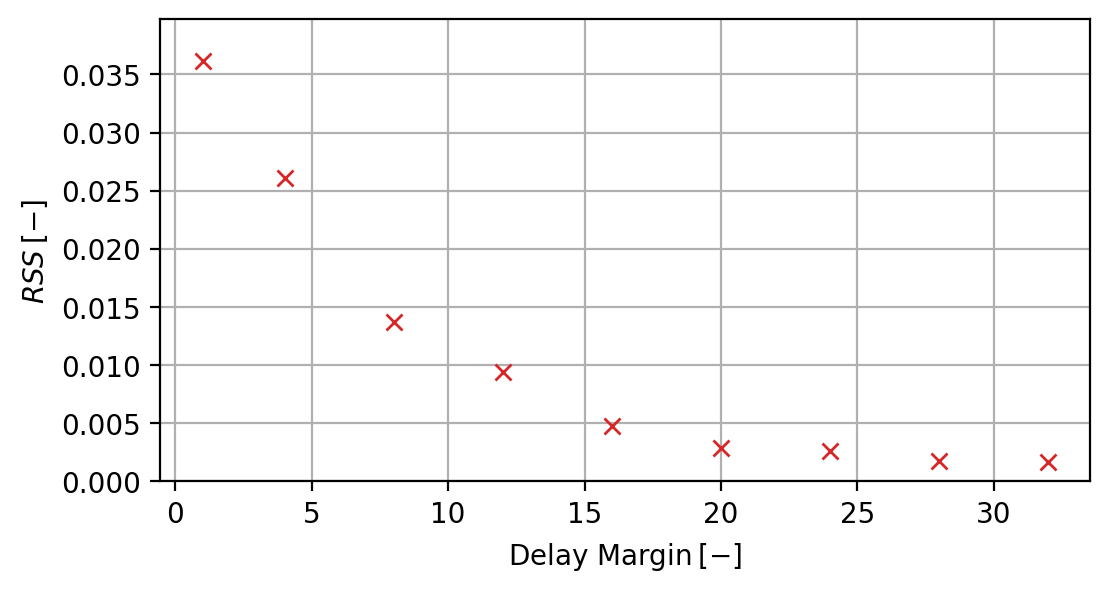}
\label{fig:margin1}
}
\subfigure[Offline performance]{
\includegraphics[width=0.80\linewidth]{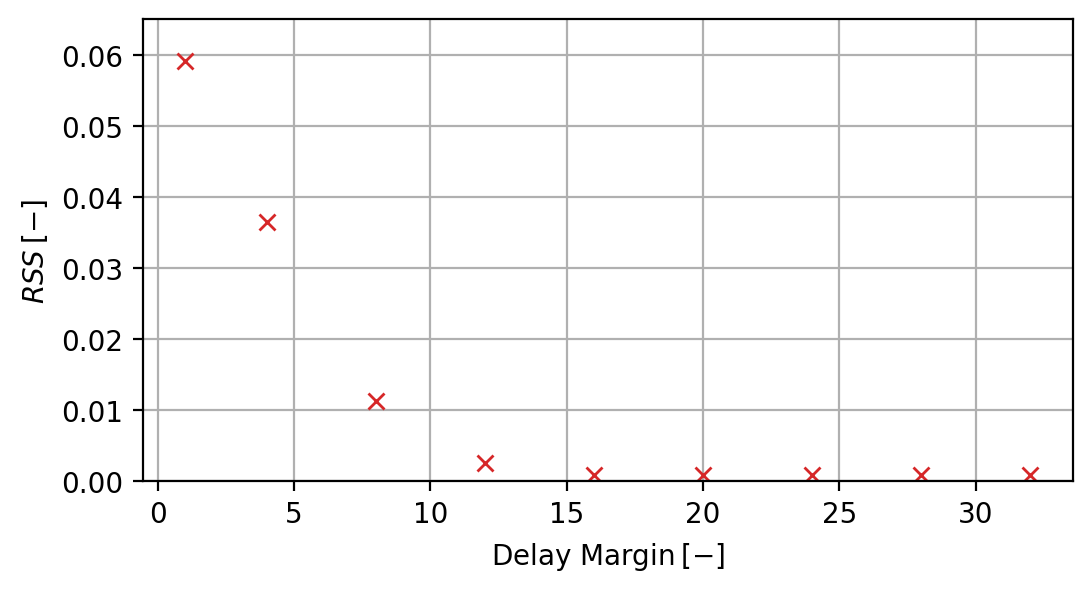}
\label{fig:margin2}
}
\caption{Performance over varying delay margin}
\label{fig:margin}
\end{figure}

As can be seen in Fig. \ref{fig:margin1} and \ref{fig:margin2}, the performance increases with increasing delay margin with quickly diminishing returns. This is because the delay margin improves the performance in two ways: it compensates for changing delays, and it provides redundancy for packet loss. If the delay varies very slightly and the packet always arrives within a few steps of the predicted arrival time, the delay margin does not need to be much larger than the tolerance window.

\subsubsection{Robustness against Packet Loss}
Another factor to examine was the robustness of the system against packet loss. For this, the packet loss in the bottleneck between $C$ and $D$ was increased in steps. Note that the packet loss was applied in both ways, meaning that a packet loss of $25\%$ means that only $75\%$ of the state packets arrive at the controller to generate control packets of which another $25\%$ would be lost on the way back to the plant for an arriving total of $56,25\%$.

\begin{figure}[htb]
\begin{center}
\includegraphics[width=0.85\linewidth]{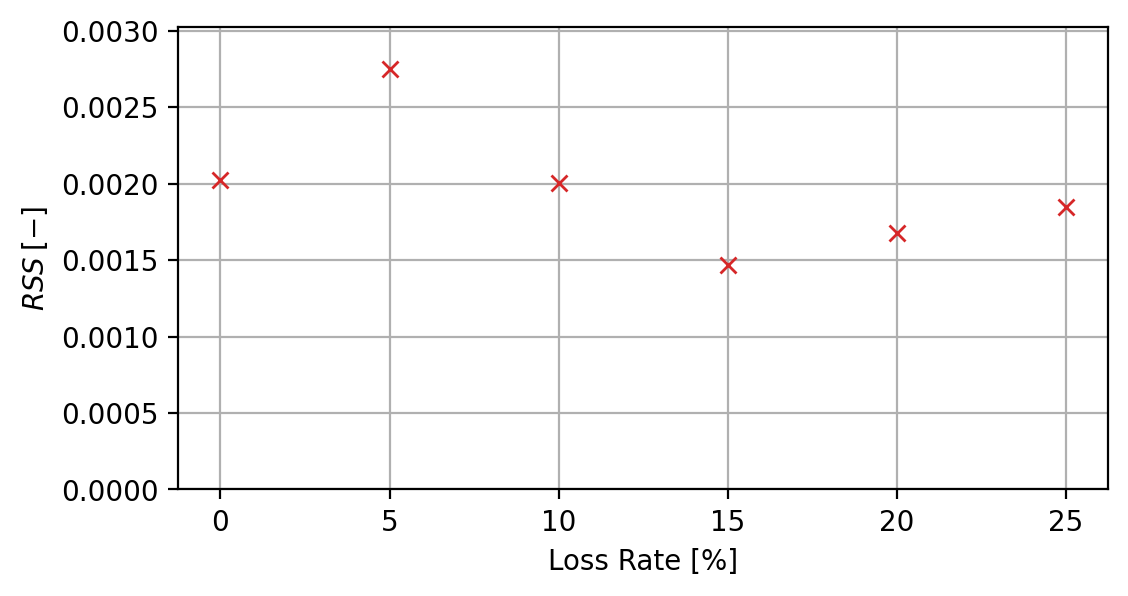}
\caption{Performance over Varying Packet Loss Rates}
\label{loss}
\end{center}
\end{figure}

Fig. \ref{loss} shows inconsistent trend, but the system is robust against packet loss. This is due to high redundancy of the system with a delay margin of $16$. The system should be able to deal with losing a majority of the packets as long as the control sequences of the remaining packets overlap.

\section{Conclusion} \label{sec4}
A networked predictive control system is implemented that can stabilize a control loop which is closed through a network by compensating for adverse effects like variable time delays and packet loss in a universal robot. The resulting controller was able to stabilise the system despite a high delay of $250 ms$ and a considerable packet loss of up to $25\%$. The offline simulations showed promising results that could be verified in real time simulations. It was shown that the predictor is able to provide useful predictions despite an uncertainty in the plant state and parameters. The introduced estimator was able to provide further improvements to the capabilities of the predictor by continually accounting for changes in the plant model. Other suggested improvements such as reusing old control values to increase the accuracy of the prediction or predicting the state of the system using an integrator instead of using matrix multiplication were shown to be effective. Efforts to cut down on excessive redundancy by introducing the delay margin were shown to have only a small negative effect on performance while significantly reducing the bandwidth needed. The remaining redundancy in the system was shown to be high enough to make it very robust against packet loss or packet reordering, which behaved very similarly in this context. Since the simulation experiments proved promising, the next logical step will be to go for full real-world experiments. The simulated plant can be replaced with the real robot arm. Routing the data can be through internet between two sites that are separated by a large geographical distance.

\bibliographystyle{IEEEtran}
\bibliography{references.bib}

\begin{thebibliography}{10}
\providecommand{\url}[1]{#1}
\csname url@samestyle\endcsname
\providecommand{\newblock}{\relax}
\providecommand{\bibinfo}[2]{#2}
\providecommand{\BIBentrySTDinterwordspacing}{\spaceskip=0pt\relax}
\providecommand{\BIBentryALTinterwordstretchfactor}{4}
\providecommand{\BIBentryALTinterwordspacing}{\spaceskip=\fontdimen2\font plus
\BIBentryALTinterwordstretchfactor\fontdimen3\font minus
  \fontdimen4\font\relax}
\providecommand{\BIBforeignlanguage}[2]{{%
\expandafter\ifx\csname l@#1\endcsname\relax
\typeout{** WARNING: IEEEtran.bst: No hyphenation pattern has been}%
\typeout{** loaded for the language `#1'. Using the pattern for}%
\typeout{** the default language instead.}%
\else
\language=\csname l@#1\endcsname
\fi
#2}}
\providecommand{\BIBdecl}{\relax}
\BIBdecl

\bibitem{Li}
\BIBentryALTinterwordspacing
M.~{Li} and Y.~{Chen}, ``Challenging research for networked control systems: A
  survey,'' \emph{Transactions of the Institute of Measurement and Control},
  vol.~41, no.~9, pp. 2400--2418, 2019. [Online]. Available:
  \url{https://doi.org/10.1177/0142331218799818}
\BIBentrySTDinterwordspacing

\bibitem{Zhang}
X.~{Zhang}, Q.~{Han}, X.~{Ge}, D.~{Ding}, L.~{Ding}, D.~{Yue}, and C.~{Peng},
  ``Networked control systems: a survey of trends and techniques,''
  \emph{IEEE/CAA Journal of Automatica Sinica}, vol.~7, no.~1, pp. 1--17, 2020.

\bibitem{halevi1988integrated}
Y.~Halevi and A.~Ray, ``Integrated communication and control systems: Part
  i—analysis,'' 1988.

\bibitem{luck1994experimental}
R.~Luck and A.~Ray, ``Experimental verification of a delay compensation
  algorithm for integrated communication and control systems,''
  \emph{International Journal of Control}, vol.~59, no.~6, pp. 1357--1372,
  1994.

\bibitem{nilsson1998real}
J.~Nilsson \emph{et~al.}, ``Real-time control systems with delays,'' 1998.

\bibitem{xi1998planning}
N.~Xi and T.-J. Tarn, ``Planning and control of internet-based teleoperation,''
  in \emph{Telemanipulator and Telepresence Technologies V}, vol. 3524.\hskip
  1em plus 0.5em minus 0.4em\relax International Society for Optics and
  Photonics, 1998, pp. 189--195.

\bibitem{liu2004networked}
G.~Liu, J.~Mu, D.~Rees \emph{et~al.}, ``Networked predictive control of systems
  with random communication delay,'' in \emph{Proceedings of UKACC
  International Conference on Control}, 2004.

\bibitem{zhang2020model}
J.~Zhang, S.~Chai, and B.~Zhang, ``{Model-Based Event-Triggered Dynamic Output
  Predictive Control of Networked Uncertain Systems with Random Delay},''
  \emph{International Journal of Systems Science}, vol.~51, no.~1, pp. 20--34,
  2020.

\bibitem{zhang2012design}
J.~Zhang, Y.~Xia, and P.~Shi, ``Design and stability analysis of networked
  predictive control systems,'' \emph{IEEE Transactions on Control Systems
  Technology}, vol.~21, no.~4, pp. 1495--1501, 2012.

\bibitem{li2018robust}
M.~Li and Y.~Chen, ``Robust adaptive sliding mode control for switched
  networked control systems with disturbance and faults,'' \emph{IEEE
  Transactions on Industrial Informatics}, vol.~15, no.~1, pp. 193--204, 2018.

\bibitem{vafamand2018networked}
N.~Vafamand, M.~H. Khooban, T.~Dragi{\v{c}}evi{\'c}, and F.~Blaabjerg,
  ``Networked fuzzy predictive control of power buffers for dynamic
  stabilization of dc microgrids,'' \emph{IEEE Transactions on Industrial
  Electronics}, vol.~66, no.~2, pp. 1356--1362, 2018.

\bibitem{pang2021active}
Z.-H. Pang, C.-G. Xia, J.~Sun, G.-P. Liu, and Q.-L. Han, ``Active
  fault-tolerant predictive control of networked systems subject to actuator
  faults and random communication constraints,'' \emph{International Journal of
  Control}, pp. 1--7, 2021.

\bibitem{chen2011stability}
D.~Chen, X.~Tang, N.~Xi, Y.~Wang, and H.~Li, ``Stability analysis for internet
  based teleoperated robot using prediction control,'' in \emph{2011 IEEE
  International Conference on Cyber Technology in Automation, Control, and
  Intelligent Systems}.\hskip 1em plus 0.5em minus 0.4em\relax IEEE, 2011, pp.
  138--143.

\bibitem{zhang2015survey}
X.-M. Zhang, Q.-L. Han, and X.~Yu, ``Survey on recent advances in networked
  control systems,'' \emph{IEEE Transactions on industrial informatics},
  vol.~12, no.~5, pp. 1740--1752, 2015.

\bibitem{mahmoud2016networked}
M.~S. Mahmoud, ``Networked control systems analysis and design: An overview,''
  \emph{Arabian journal for science and engineering}, vol.~41, no.~3, pp.
  711--758, 2016.

\bibitem{pollak2020measurement}
M.~Poll{\'a}k, M.~Ko{\v{c}}i{\v{s}}ko, D.~Pauli{\v{s}}in, and P.~Baron,
  ``Measurement of unidirectional pose accuracy and repeatability of the
  collaborative robot ur5,'' \emph{Advances in Mechanical Engineering},
  vol.~12, no.~12, p. 1687814020972893, 2020.

\bibitem{shaiju2008formulas}
A.~Shaiju and I.~R. Petersen, ``{Formulas for Discrete Time LQR, LQG, LEQG and
  Minimax LQG Optimal Control Problems},'' \emph{IFAC Proceedings Volumes},
  vol.~41, no.~2, pp. 8773--8778, 2008.

\bibitem{fuller2021python}
S.~Fuller, B.~Greiner, J.~Moore, R.~Murray, R.~van Paassen, and R.~Yorke, ``The
  python control systems library (python-control),'' in \emph{2021 60th IEEE
  Conference on Decision and Control (CDC)}.\hskip 1em plus 0.5em minus
  0.4em\relax IEEE, 2021, pp. 4875--4881.

\end{thebibliography}

\end{document}